# Mobile Robot Motion Control Using a Combination of Fuzzy Logic Method and Kinematic Model

Anh-Tu Nguyen[1(✉)] and Cong-Thanh Vu[2]

[1] Faculty of Mechanical Engineering, Hanoi University of Industry, Hanoi 159999, Vietnam
`tuna@haui.edu.vn`
[2] Faculty on Mechanical Engineering, University of Economics - Technology for Industries, Hanoi 159999, Vietnam
`vcthanh@uneti.edu.vn`

**Abstract.** Mobile robots have been widely used in various aspects of human life. When a robot moves between different positions in the working area to perform the task, controlling motion to follow a pre-defined path is the primary task of a mobile robot. Furthermore, the robot must remain at its desired speed to cooperate with other agents. This paper presents a development of a motion controller, in which the fuzzy logic method is combined with a kinematic model of a differential drive robot. The simulation results are compared well with experimental results indicate that the method is effective and applicable for actual mobile robots.

## 1 Introduction

Motion control for the mobile robot has been an interesting topic and received the attention of numerous researchers. The objective of robot motion control is to follow pre-defined trajectories represented by its position, velocity, or acceleration profile as a function of time. The trajectory is normally divided into basic segments such as straight lines, circle pieces, or curves. The control problem is thus to precompute a smooth trajectory-based online and circle segments which drive the robot from the start point to the goal point [1]. Conventional control methods are mainly developed using the kinematic and dynamics governing equation system. The kinematic control system is represented by kinematic equations and the control inputs are the pure velocity [2, 3]. This method has shown several disadvantages because the robot will not automatically adapt to the change of the environment or the unsmooth transition between trajectory segments. This results in a discontinuous acceleration of the motion. To solve the limitations regarding the kinematic control method, the dynamics control has been developed, in which both torque and force are used as the input parameters and the dynamics model is necessary for designing the controller [4–6]. However, it is almost impossible to build a perfect dynamics model of a robot in real applications. As such, fuzzy logic control methods were introduced to cope with parameter uncertainties, therefore, improve the motion control quality [7–10]. The fuzzy logic controller is normally applied in combination with the PID controller. The fuzzy logic controller detects the PID parameter to adapt to the working environment. This method help robots tract the pre-defined path better [8]. The simulation results showed a better control quality compared with the traditional control method [9]. Another application of fuzzy logic control that should be cited here is the method based on Z-number for tracking trajectory. This approach helps coding works become easier and deal with missing observations [10]. Recently, with the development of technologies in hardware such as vision sensors and artificial intelligence. Robots become smarter and can autonomously perform complex tasks, cooperate with other agents with high precision, high speed, and stable safety [11].

This paper presents an approach for controlling mobile robot motion, in which the fuzzy method is incorporated with the kinematic model to manage the linear velocity while following the trajectory. The proposed method then is applied in a differential drive mobile robot for verification. The good comparison between simulation and experiment proves that the robot can follow the pre-defined path and manage desired linear velocity while remains an acceptable angular velocity.

## 2  Kinematic Model

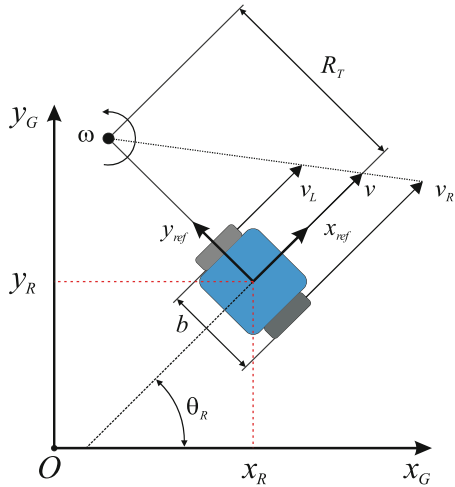

**Fig. 1.** The demonstration of the robot concerning reference and global coordinate system.

This section addresses the relationship between different geometrical factors and kinematic parameters of a robot model. Because the reference coordinate system ($Ox_{ref} y_{ref}$) is fixed at the center of the robot, therefore the position ($x_R$, $y_R$) and orientation ($\theta_R$) are the relative motion between this coordinate system and the global coordinate system ($O x_G y_G$) as illustrated in Fig. 1.

The tangential velocity of two wheels ($v_L$, $v_R$) are in the relation with the agular velocity ($\omega_L$, $\omega_R$) as following:

$$v_R = \omega_R R \tag{1}$$

$$v_L = \omega_L R \tag{2}$$

Therefore, the robot linear velocity can be determined as:

$$v = \frac{v_R + v_L}{2} \tag{3}$$

$$v = \frac{(\omega_R + \omega_L)R}{2} \tag{4}$$

Supposing the distance between two wheels is $b$, the angular velocity of the robot is:

$$\omega = \frac{v_R - v_L}{b} \tag{5}$$

$$\omega = \frac{(\omega_R - \omega_L)R}{b} \tag{6}$$

From the Eqs. (4) and (6), the angular velocities of each wheel in the relation with $v$ and $\omega$ can be expressed:

$$\omega_R = \frac{v + b\omega}{R} \tag{7}$$

$$\omega_L = \frac{v - b\omega}{R} \tag{8}$$

## 3 Fuzzy Control Method

To enhance the robot operation, this section presents an approach in which the controller is designed by using the kinematic model and fuzzy logic method. The control system is shown in Fig. 2, in which inputs of the fuzzy block involve the heading angle error ($\varepsilon_e$) and linear velocity ($v$) of the robot. After defuzzification, the output of the fuzzy controller, the angular velocity ($\omega$), is used along with the linear velocity for calculating the angular velocity of each robot's wheel via inverse kinematic equations. This helps detect appropriate angular velocity to remain the linear velocity stable while still following the path.

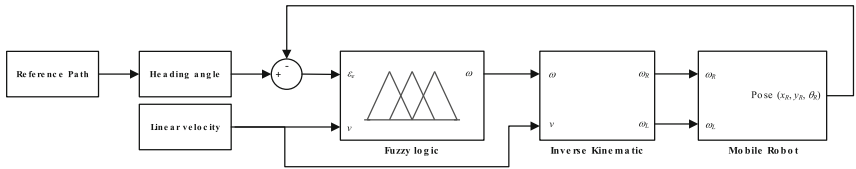

**Fig. 2.** The block diagram of the control system.

The fuzzy controller involves three main units as in Fig. 3: fuzzification, fuzzy inference, and defuzzification. Figure 4 and Table 1 show the input membership functions and the fuzzy control outputs. If the heading angle error is small, the output angular velocity is low. Otherwise, it is high. This output angular velocity will also depend on the setup linear velocity. Table 2 describes the definition of the fuzzy rules.

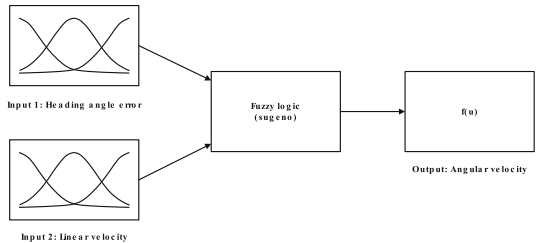

**Fig. 3.** Configuration of fuzzy control.

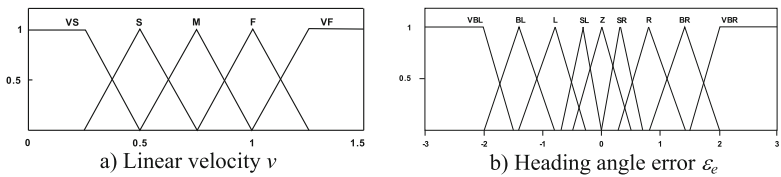

a) Linear velocity $v$     b) Heading angle error $\varepsilon_e$

**Fig. 4.** Membership function plotting of $v$ and $\varepsilon_e$.

**Table 1.** Membership function plots of output variable angular velocity.

| Information | Name | Value |
| --- | --- | --- |
| Very big left deviation | VBL | − 5 |
| Big left deviation | BL | − 3 |
| Left deviation | L | − 1.75 |
| Small left deviation | SL | − 0.5 |
| Zero deviation | Z | 0 |
| Small right | SR | 0.5 |
| Right deviation | R | 1.75 |

(*continued*)

Table 1. (*continued*)

| Information | Name | Value |
| --- | --- | --- |
| Big right deviation | BR | 3 |
| Very big right deviation | VBR | 5 |

Table 2. Fuzzy rules for motion control.

| Heading angle error \ Linear Velocity | VS | S | M | F | VF |
| --- | --- | --- | --- | --- | --- |
| VBL | BR | VBR | VBR | VBR | VBL |
| BL | BR | BR | VBR | VBR | VBR |
| L | R | R | BR | BR | BR |
| SL | SR | SR | R | R | R |
| Z | Z | Z | Z | Z | Z |
| SR | SL | SL | L | L | L |
| R | L | L | BL | BL | BL |
| BR | BL | BL | VBL | VBL | VBL |
| VBR | BL | VBL | VBL | VBL | VBL |

## 4   Results and Discussion

To examine the effectiveness and feasibility of the combination of kinematic model and fuzzy logic control for trajectory tracking of the differential wheeled mobile robot. The control system is demonstrated in Fig. 5, including a laptop that takes the role of center processing, a microcontroller, STM32F407VET6, receives and transfers signals among the laptop, drivers, and sensor system. The active wheels were driven using two DC motors. The sensor system is designed using both relative and absolute measurement methods, encoders, a digital compass IMU, and a laser scanner NAV245.

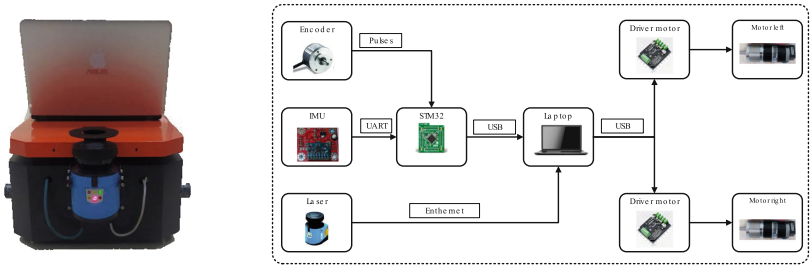

**Fig. 5.** The hardware system of the experimental robot

The trajectory is designed to include distinct profiles: a straight line along an axis, a circle, and a bias line with a total distance of 60 m. The simulation results are validated by experiments. As mentioned above, this method control robot to follow the path while remains the desired speed. Therefore, the velocity is varied to estimate the repose of the fuzzy controller.

The robot's simulation and experimental performance of trajectory following are shown in Fig. 6 in the case of $v = 0.6$ m/s. The results show that the robot can track the reference path over a long distance with different profiles. Figure 7 and Fig. 8 illustrate the specific position and orientation variation among the consideration case studies. When the robot moves in the x-direction, the error in the y-axis is pretty small while the error in the x-axis fluctuates stronger. The inversion in the error occurs when the robot runs along the y-direction. In the bias path, the position derivation in both two-axis is similar. Because the present method prioritizes to control robot's speed, therefore the simulation result of linear velocity is closed the reference value. However, there is a small oscillation in the experimental results (Fig. 9). This is understandable because the experimental performance is affected by different factors. The specific angular velocity of the robot is demonstrated in Fig. 10. The simulation and experimental results are compared well in almost the motion. Nevertheless, when the robot runs in the transition path, the head angle fluctuates stronger than the linear velocity. The estimation of the standard error is evaluated by Root Mean Square Error (RMSE) method in position, the linear and angular velocities after 30 trials are addressed in Table 3. In all consideration case studies, the error increases slightly in all parameters, however, the average error of the robot's linear velocity is less than 9 mm while the other parameters remain under the acceptable values for the robot activities.

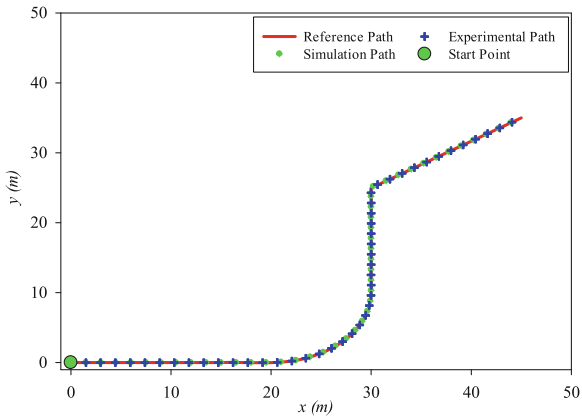

**Fig. 6.** Trajectory tracking at $v = 0.6$ m/s.

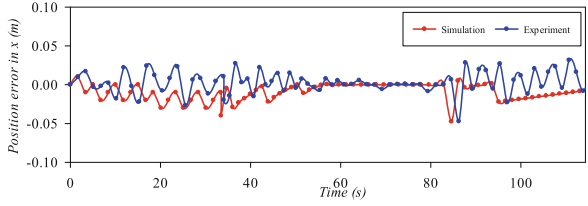

**Fig. 7.** The specific position derivation in x direction at $v = 0.6$ m/s.

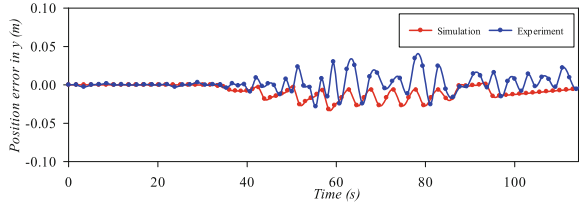

**Fig. 8.** The specific position derivation in y direction at $v = 0.6$ m/s.

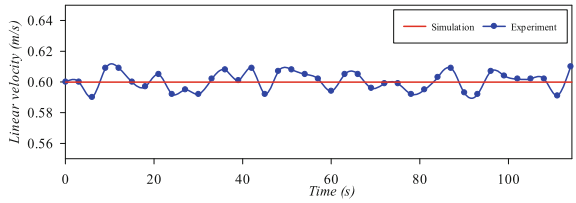

**Fig. 9.** The simulation results of linear and angular velocities at $v = 0.6$ m/s.

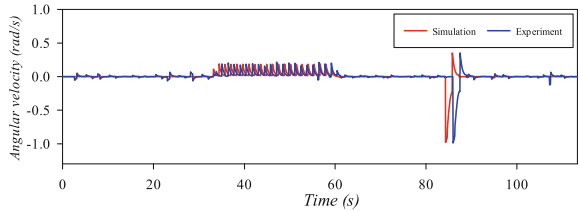

**Fig. 10.** The experimental results of linear and angular velocities at $v = 0.6$ m/s.

Table 3. The error estimation at different velocity.

| No | Parameter | $v = 0.3$ m/s | | $v = 0.6$ m/s | | $v = 1.2$ m/s | |
|---|---|---|---|---|---|---|---|
| | | Simulation | Experiment | Simulation | Experiment | Simulation | Experiment |
| 1 | Position error in x-axis (m) | 0.006 | 0.009 | 0.014 | 0.015 | 0.019 | 0.025 |
| 2 | Position error in y-axis (m) | 0.006 | 0.008 | 0.012 | 0.013 | 0.013 | 0.017 |
| 3 | Linear velocity error (m/s) | 0 | 0.006 | 0 | 0.007 | 0 | 0.009 |
| 3 | Angular velocity error (rad/s) | 0.05 | 0.064 | 0.082 | 0.085 | 0.11 | 0.124 |

## 5  Conclusion

This paper presents an approach of fuzzy logic along with the kinematic model to develop a controller for a wheeled mobile robot. Since the robot's pose plays an important role in doing tasks and is determined by angular and linear velocities, therefore, the angular error and the reference velocity are both used as inputs of fuzzy logic controller. The experiments indicates that the robot can successfully follow the trajectories and operate in desired linear velocity while remains acceptable angular velocity.